\documentclass[letterpaper]{article} 

\usepackage{aaai19}  
\usepackage[usenames,dvipsnames,dvinames]{xcolor}
\usepackage{graphicx}
\PassOptionsToPackage{usenames,dvipsnames,dvinames}{xcolor}
\usepackage[utf8]{inputenc}
\usepackage{wrapfig}
\usepackage{lipsum}
\usepackage{microtype}
\usepackage{amsthm}
\usepackage{amsmath}
\newtheorem{theorem}{Theorem}
\newtheorem{corollary}{Corollary}[theorem]

\providecommand{\doarxiv}{true}
\usepackage{ifthen}
\newboolean{isarxiv}
\setboolean{isarxiv}{\doarxiv} 
\ifthenelse{\boolean{isarxiv}}{%
\newcommand{\arxiv}[1]{#1}
\newcommand{\notarxiv}[1]{}
}{
\newcommand{\arxiv}[1]{}
\newcommand{\notarxiv}[1]{#1}
}
\newcommand{\arxivalt}[2]{\ifthenelse{\boolean{isarxiv}}{#1}{#2}}
\newcommand{\arxivaltr}[2]{\ifthenelse{\boolean{isarxiv}}{#2}{#1}}
\newcommand{\narxiv}[1]{\notarxiv{#1}}

\narxiv{}
\usepackage[bookmarks, colorlinks=true, citecolor=Violet,linkcolor=Mahogany,urlcolor=blue]{hyperref}

\begin{document}
\title{A Unified Batch Online Learning Framework for Click Prediction}
\author{Rishabh Iyer, Nimit Acharya, Tanuja Bompada, Denis Charles, Eren Manavoglu \\
Bing Ads, Microsoft \\
One Microsoft Way, Redmond, WA, USA}
\maketitle
\begin{abstract}
We present a unified framework for Batch Online Learning (OL) for Click Prediction in Search Advertisement. Machine Learning models once deployed, show non-trivial accuracy and calibration degradation over time due to model staleness. It is therefore necessary to regularly update models, and do so automatically. This paper presents two paradigms of Batch Online Learning, one which incrementally updates the model parameters via an early stopping mechanism, and another which does so through a proximal regularization. We argue how both these schemes naturally trade-off between old and new data. We then theoretically and empirically show that these two seemingly different schemes are closely related. Through extensive experiments, we demonstrate the utility of of our OL framework; how the two OL schemes relate to each other and how they trade-off between the new and historical data. We then compare batch OL to full model retrains, and show how online learning is more robust to data issues. We also demonstrate the long term impact of Online Learning, the role of the initial Models in OL, the impact of delays in the update, and finally conclude with some implementation details and challenges in deploying a real world online learning system in production. While this paper mostly focuses on application of click prediction for search advertisement, we hope that the lessons learned here can be carried over to other problem domains. 
\end{abstract}

\section{Introduction}
Click prediction is an important and central component in any online advertisement system. Predicting the probability of clicks and click through rate is central in sponsored search advertising and display advertising, and several downstream systems including our auction mechanism rely on being able to predict the probability of click accurately and reliably. Most click prediction systems are modeled via the standard machine learning classification framework. We design features relevant to the user, ads and query, with the goal of predicting if a given user, will click a given ad for a given search query. A training data period is selected, and the click prediction model is trained and validated. Machine learning scientists then analyze the model performance offline, and if things look good, deploy the models in production.

\begin{figure}
\includegraphics[width = 0.5\textwidth]{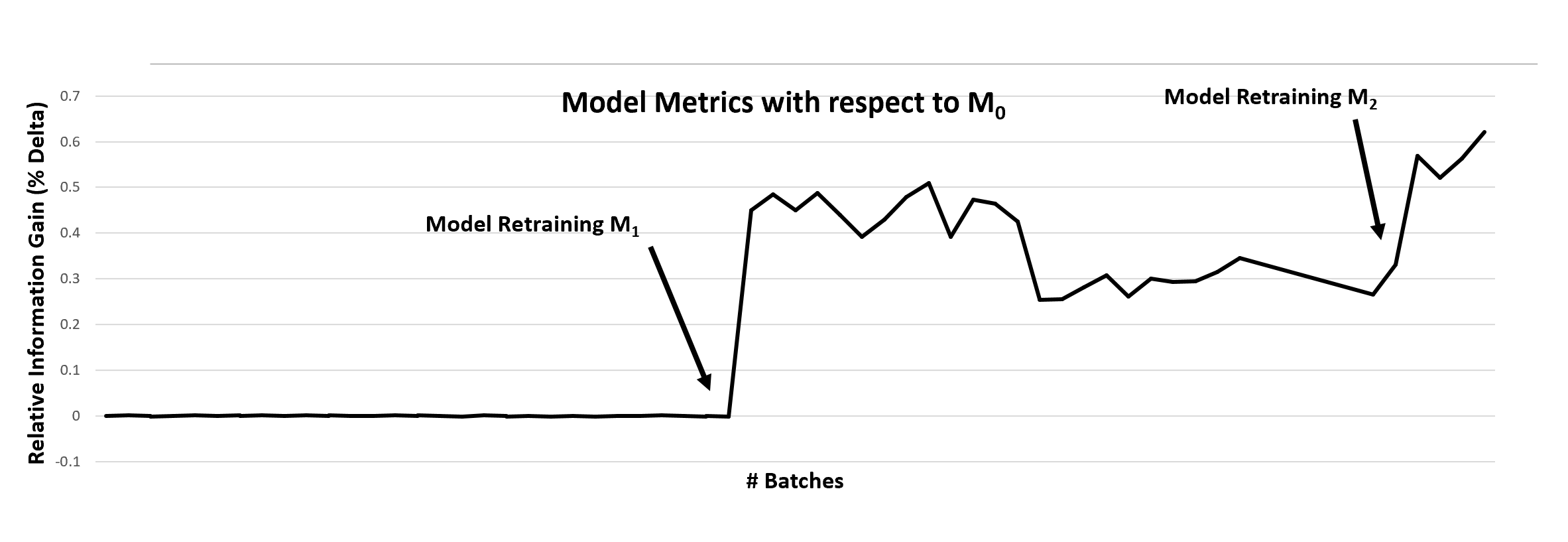}
\caption{Effect of Staleness on Model Metrics and gains with retraining}
\end{figure}

The main caveat of this, is that the models get stale quickly and the performance degrades over time. The distribution of users, queries and ads change over time and correspondingly models are evaluated on a different data distribution compared to what they have learnt from. For this reason, we need to retrain model periodically. Figure 1 demonstrates that we can achieve a significant gain in model metrics after retraining the model in just a month.

This paper addresses the issue of model staleness via online learning. We investigate a unified model adaptation framework, where the models continuously and gradually adapt over time, to learn the distribution changes. A natural alternative is to just run an automated experimental pipeline which continuously retrains the model from scratch. We not only show that gradual learning outperforms this, but we also show that via continuous adaptation, we are able to control how much the model adapts and ensure the model does not overfit to a single distribution. For example, we know that the distribution of queries, ads and users are quite different on Labor Day, compared to other week days. Training a model from scratch on data containing Labor Day can be problematic since it will learn a potentially different input distribution, that is not seen on other days. This problem gets intensified when there is data or system corruption issues. Our comprehensive evaluation shows that our batch online learning framework via gradual and continuous model adaption, not only improves performance, but is also more reliable and safe compared to automatic model retraining.

\subsection{Related Work}
The problem of online learning has been heavily studied in the theoretical machine learning community. Most of this work has revolved regret minimization, where the regret of an online learning algorithm is defined as the gap in performance of the online algorithm compared to the solution obtained from an offline algorithm, which has access to all the data in hindsight~\cite{zinkevich2003online,shalev2012online,shalev2007online}. Since several machine learning problems naturally involve solving convex optimization problems, online machine learning can naturally be posed as online convex optimization. This is the case with all linear models like logistic regression, SVMs etc. Zinkevich~\cite{zinkevich2003online} was one of the first papers to study online gradient descent for online learning and show that OGD enjoys low regret. Following this seminal work, several papers have extended upon this paradigm (see~\cite{shalev2012online,shalev2007online} for a survey).

While most literature around online learning has focused on proving theoretical bounds, a few of these have been proved successful in real world problems. \cite{mcmahan2013ad,mcmahan2011follow} proposed a Follow the Regularized Leader Scheme (FTRL) for online learning on a logistic regression model. \arxiv{Their problem setting consists of a sparse feature set (with more than a million features) with a L1 logistic regression model. The authors argue how their framework naturally handles both L1 and L2 regularization, and in the case of L2 regularization, boils down to online Gradient Descent.} The authors provide extensive empirical validation of their framework and some hints into the deployment of such a large scale system for serving ads at Google. Following this \cite{he2014practical} from Facebook provide a framework of Online Learning with a combined Decision Tree and Logistic Regression Model. Similar to~\cite{mcmahan2013ad}, the authors go into a lot of details into the deployment of a real world online learning system in production. \cite{ciaramita2008online} propose an online learning click prediction system on multi-layer nueral networks. Another large scale Online Learning system for Click Prediction was proposed by~\cite{graepel2010web}, where they propose Online Learning system on Bayesian Probit Regression Models, and provide compelling details into deploying such a system in practice. Similarly~\cite{liu2017pbodl} describes the click prediction system at Tencent where they use a Bayesian Online Learning scheme similar to~\cite{graepel2010web}. \cite{cheng2010personalized} investigate the role of personalization in click prediction systems. \cite{mcmahan2014delay} investigate a distributed online learning framework for large scale click prediction problems.

Beyond  Click Prediction in Search Advertisement, Online Learning schemes have been used in other scenarios as well. \cite{chapelle2011empirical,chapelle2015simple} propose an Thompson Sampling based contextual bandit scheme for Display advertisement. They propose a Proximal Update Algorithm similar to the one discussed in this paper. Similarly, \cite{kirkpatrick2017overcoming} look into the problem of learning from a new domain, while simultaneously not forgetting about the previous domain. Similarly \cite{ma2009identifying} investigate online learning for identifying suspicious URLS.

\subsection{Our Contributions}
The following are our main contributions.
\begin{itemize}
\item This paper studies two different views of Online Learning: One which performs iterative training (like Online Gradient Descent, FTRL etc.) with early stopping, and another, which minimizes at every round, a proximal regularized objective function (which ensures the current solution does not move too much from the previous solution). Both approaches provide tradeoff between historical and new data, via the learning rate and number of iterations in the early stopping, and the regularization parameter in the proximal scheme.
\item We empirically and theoretically show that both these paradigms are closely related to each other. In particular, we show that with a right choice of these parameters (learning rate, number of iterations and proximal regularization), the two OL paradigms achieve very similar solutions. 
\item We next prove the benefit of incremental learning schemes, by showing how this can substantially improve upon simple retraining of models. We argue how online learning not only ensures automatic model updates, but also can improve upon model metrics because of the fact it  retains a larger history. Moreover, we also show how it is much more robust to data corruption and other distributional changes compared to simple model retrainings.
\item We then look into several important challenges of production systems and study the effect of online learning with data delays, how different initializations affect the performance of OL, and we conclude by discussing engineering issues in deploying such a model in production systems serving Search Ads to Hundreds of Millions of Users. 
\end{itemize}

\section{System Overview}
\begin{wrapfigure}{L}{0.25\textwidth}
 \centering
\includegraphics[width = 0.22\textwidth]{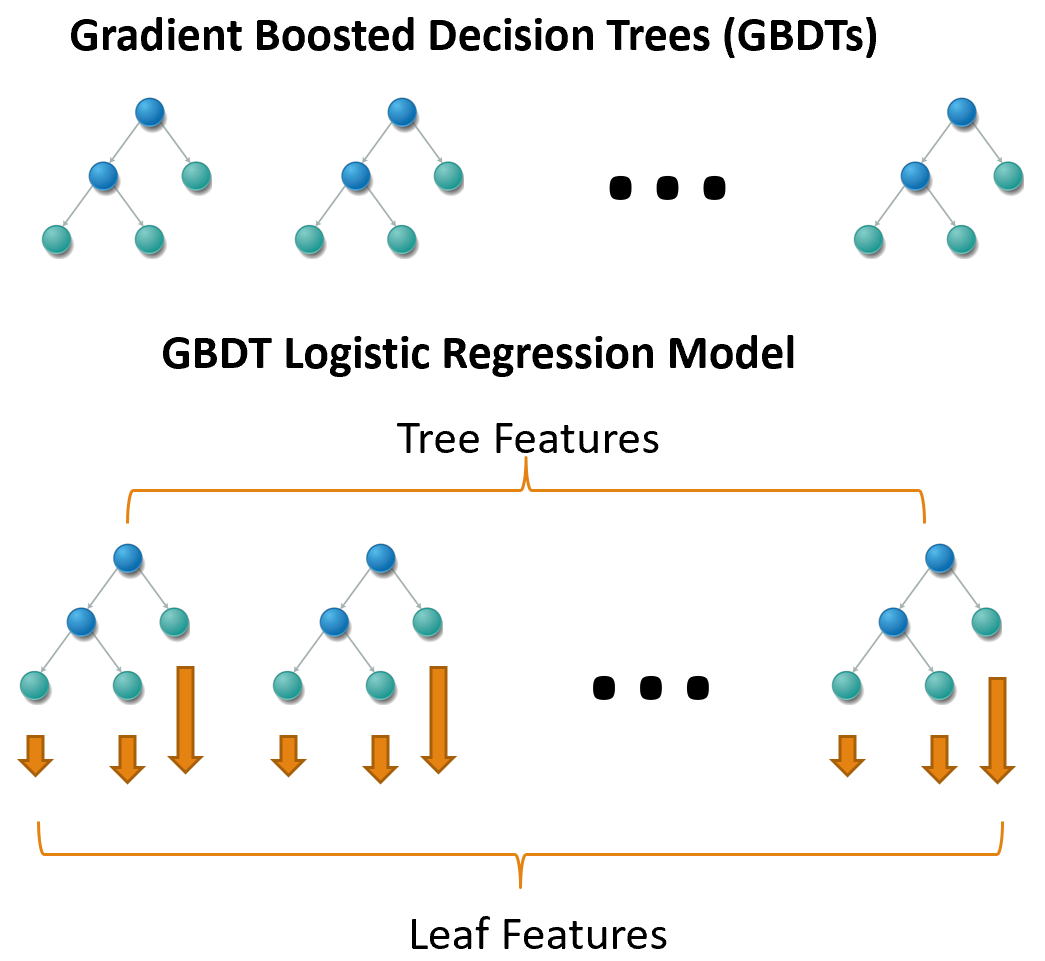}
\caption{GBDT as a Feature Extractor for Linear Models}
\end{wrapfigure}
In this section, we go over our modeling framework, features, evaluation metrics and our system overview. Given a user, ad and query, our task is to accurately predict the probability that the user will click on this ad. It is not just important to rank the ads correctly, but the resulting probability must be calibrated (in that the predicted probability must match the true click through rate). For this reason, we shall compare both the Area under the curve (AUC), which measures the ranking of the ads and the Relative Information Gain (RIG) metrics which measures the calibration. The RIG of a Model $M$ can be defined as 
\begin{equation}
RIG_M = \frac{LogLoss_M - LogLoss_{CTR}}{LogLoss_{CTR}}
\end{equation}
where $LogLoss_{CTR}$ is the LogLoss of the empirical CTR of the data. Since $LogLoss_{CTR}$ is a constant, RIG is proportional to the LogLoss of the Model. 

Next, we go over the features for our problem. Our features include Ad, Query and User features. Ad features include Ad Title, Ad id, Decoration information etc. Query features include query category, query text etc. User features include IP address, Browser, Location, age/gender information etc. We encode our features as Counting features~\cite{ling2017model}, representing the click through rate for that feature. We resort to two of the most popular choices of supervised learning techniques, namely gradient boosted decision trees and Neural Networks. Both these techniques outperform other non-linear and generalized linear models on our data. To incrementally train models over time however, it is more natural to do so over generalized linear models. We achieve best of both worlds, by training a generalized linear model over features extracted from non-linear models. For example, we can extract tree and leaf value features from a GBDT (shown in the Figure 2), and train a Logistic Regression model on these features. This can also be done if we use a Neural Network as the feature extractor, and if we extract features, say from the last layer. In this paper, we shall focus on Online Learning over a Logistic Regression Model using features extracted from a fixed GBDT model.

\section{Our Online Learning Framework}
In Click Prediction systems, we get near instant feedback from users based on whether they click on an ad or not. Assume we have a Base LR Model $M_{\mathcal S}$, trained on a given dataset $\mathcal S$ (say, for example, one week of data). Once a user searches for a query on a search engine, the system sees a feature vector $x_t \in \mathbf{R}^d$. Using the Model currently in production, the system then predicts the probability of click $p_t$. The auction then ranks ads based on the pClick, bid and other factors, finally creating a set of ads which are shown to the user. We then receive the feedback $y_t$ whether the user clicks or not. This data is then collected in batches. 
Denote $\mathcal B_1, \mathcal B_2, \cdots, \mathcal B_n$ as the different batches of data (each batch, is for example, a day of data or four hours). The predictions made in batch $\mathcal B_i$ are made using the Model from the previous batch -- i.e. $M_{i-1} = M_{\mathcal B_{i-1}}$. 

The most important piece of this story is how do we update the model. A critical challenge here is to be able to learn from the incremental data coming in, and yet, not forget what was learned in the past. In the below sections, we describe two schemes of incremental updates of the models.

\subsection{Early Stopping Incremental Learning}
The first scheme, is what we call {\em early stopping} scheme, abbreviated as {\sc ES}. We initialize the model with the base model $M_{\mathcal S}$. At round $i$, we initialize the incremental learning algorithm $\mbox{Alg}$, with the model from the previous round $M_{i-1}$, and limit the number of passes on the data to be $k$. We denote this by, 
\begin{equation} \label{esupdate}
M_i = \mbox{Alg}(M_{i-1}, \mathcal B_i, k).
\end{equation}

\arxiv{\noindent \textbf{LBFGS/TRON:} One example of $\mbox{Alg}$ is LBFGS~\cite{liu1989limited}. Limited-memory BFGS (LBFGS) algorithm belongs to a family of quasi-Newton methods which approximates the BFGS algorithm with a limited memory. 
The BFGS algorithm itself is an iterative technique, where the Hessian matrix is updated at every iteration using the past gradient evaluations. BFGS requires storing the dense $n \times n$ approximation of the Inverse Hessian Matrix, while L-BFGS just stores the past $m$ updates of the positions and gradients and uses them for the updates. In practice, $m$ is chosen around $10$ – $30$. Another example of a similar algorithm is a Trust Region Newton~\cite{lin2007trust}.
}

\noindent \textbf{OGD/SGD/GD: } \narxiv{One}\arxiv{Another} choice of $\mbox{Alg}$ is Online Gradient Descent~\cite{zinkevich2003online} or Stochastic Gradient Descent~\cite{bottou2018optimization}. This is akin to a Gradient descent scheme, except that the (stochastic) gradient is computed based on a single example or a minibatch, rather than using the entire batch. There are two flavors of this, either using a fixed learning rate, a decaying learning rate or an adaptive learning rate (as in AdaGrad~
\cite{duchi2011adaptive}). 
In this paper, we focus on the simplest version of fixed learning rates for SGD or GD. The main hyper-parameters under consideration for Early Stopping algorithms is $k$ and the learning rate, which determines the tradeoff between the new data and history. Having too large a $k$, implies that we overfit to the distribution in the current batch, thereby generalizing poorly to the next batch. Having a small $k$ implies that we might learn the data changes too slowly. We shall demonstrate the interplay between these quantities in detail in our experiments. Similarly, a large learning rate can cause the incremental learning to diverge and a small learning rate could mean slow learning. One can also have a per coordinate learning rate~\cite{mcmahan2013ad,he2014practical}. One way to define a per coordinate rate, is to set $\alpha^i_k = 1/n^i_k$, where $n^i_k$ is the total number of times feature $i$ is seen till round $k$~\cite{he2014practical}.\\

\noindent \textbf{FTRL: } Follow The Regularized Leader (FTRL)~\cite{mcmahan2013ad} can be seen as another instance of this paradigm. In the case of L2 regularization, FTRL updates are equivalent to the one from OGD. 

\subsection{Proximal Regularization based Incremental Learning}
Given a Batch $\mathcal B_i = \{(x^i_1, y^i_1), \cdots, (x^i_l, y^i_l)\}$, the {\em proximal} based incremental learning scheme, abbreviated as {\sc Prox}, minimizes the following objective function:
\begin{equation} \label{proxl2}
G(w) = \sum_{j = 1}^l L(w, x^i_j, y^i_j) + \frac{\lambda}{2} ||w - w^{i-1}||^2
\end{equation}
This formulation ensures, we minimize the objective function on the current batch, while still not moving too much away from the previous solution. Here, $\lambda$ is a tradeoff between the new data and the history. If $\lambda$ is too small, we overfit completely to the current data (similar to a large $k$ in the early stopping scheme). Similarly if $\lambda$ is too large, we will not move much from the initial model. 

In Equation~\ref{proxl2}, each coordinate has the same weight $\lambda$. Often, however, we want some of the coordinates to move less compared to other coordinates. For example, coordinates that have covered many training examples in the recent history can have a higher penalty for change compared to parameters covering relatively fewer number of examples. The Proximal update equation is the same as Equation~\ref{proxl2}, except that we have a per coordinate regularization $\lambda_r$. 

\begin{equation} \label{proxewc}
G(w) = \sum_{j = 1}^l L(w, x^i_j, y^i_j) + \sum_{r = 1}^d \lambda_r (w_r - w_r^{i-1})^2
\end{equation}
where $w_r$ is the $r$th coordinate of the weight vector. This looks similar to the per coordinate learning rate in an Online Gradient Descent scheme above. One way of setting the per coordinate regularization parameter is the diagonal of the Fisher Information of the data~\cite{kirkpatrick2017overcoming}. This scheme is called Elastic Weight Consolidation in ~\cite{kirkpatrick2017overcoming}. This comes naturally as an approximation of the Posterior of the weights, which contains information of the parameters important to the historical data. In the case of Logistic Regression, this is exactly the Double derivative of the Log Likelihood Function. Incidentally, this scheme was also proposed as an online learning scheme with Thompson Sampling for Click Prediction Problems~\cite{chapelle2011empirical}. 

The individual optimization problems of the L2 Proximal Update (Equation~\ref{proxl2}) and the Per coordinate one (EWC) are convex optimization problems and can be optimized via methods like LBFGS~\cite{liu1989limited} or Trust Region Newton~\cite{lin2007trust}. 

\subsection{Relationship between Early Stopping and Proximal Update Schemes}
We next study the relationship between the early stopping algorithms and the proximal update scheme. For simplicity, we shall analyze the case when the {\sc ES} algorithm is Gradient Descent with a fixed Learning rate.  The learning rate $\alpha$ and the number of iterations $k$ for {\sc ES},  and the regularization parameter $\lambda$ for {\sc Prox} determine the trade-off and performance. We show here that there is a close relationship between the two. 

Assume we initialize {\sc ES} and {\sc Prox} with $w_0$, i.e. {\sc Prox} minimizes $G(w) = F(w) + \lambda/2 ||w - w_0||^2$, where $F(w) = \sum_{j = 1}^l L(w, x_j, y_j)$. Denote $w_1, \cdots, w_k$ as the weights obtained via an ES scheme. For this analysis, we assume we use gradient descent.  We then show the following result.
\begin{theorem}
Denote $w^*$ as the optimal solution of the {\sc Prox} objective function $G$ with regularization parameter $\lambda$. Denote by $w_k$ the solution obtained by running {\sc ES} on $F$ with a learning rate $\alpha$ for $k$ iterations. If $\lambda, \alpha$ and $k$ satisfy $\alpha \lambda k = 1$, then the solution $w_k$ satisfies,
\begin{align}
|G(w_k) - G(w^*)| \leq \epsilon (k - 1) ||w_k - w^*|| 
\end{align}
where $\epsilon = \max_i || \nabla F(w_i) - \nabla F(w_{i-1}) ||$.
\end{theorem}
The above theorem shows that as long as the gradients of the loss function $F$ do not change much from iteration $i$ to $i+1$ in the early stopping scheme, $w_k$ is close to the optimal solution of {\sc Prox} provided the parameters satisfy $\alpha \lambda k = 1$. The proof of this result is in the \arxiv{Appendix.}\narxiv{extended version.}

Theorem 1 can be extended to show a relationship between a per co-ordinate learning rate and per co-ordinate regularization. In particular, a per co-ordinate learning rate $\alpha_1, \cdots, \alpha_m$ is closely related to a per co-ordinate regularization $\lambda_1, \cdots, \lambda_m$ if $\forall r, \lambda_r \alpha_r k = 1$. 
\begin{corollary}
Denote $w^*$ as the optimal solution of the {\sc Prox} objective function $G$ with per-coordinate regularization $\lambda_r$. Denote by $w_k$ the solution obtained by running {\sc ES} on $F$ with a per-coordinate learning rate $\alpha_r$ for $k$ iterations. If $\lambda_r, \alpha_r, \forall r$ and $k$ satisfy $\alpha_r \lambda_r k = 1, \forall r$, then the solution $w_k$ satisfies,
\begin{align}
|G(w_k) - G(w^*)| \leq \epsilon (k - 1) ||w_k - w^*|| 
\end{align}
where $\epsilon = \max_i || \nabla F(w_i) - \nabla F(w_{i-1}) ||$.
\end{corollary}
We make several important remarks about Theorem 1. Firstly, as noted earlier, $w_k$ (obtained via $k$ rounds of the ES scheme) is close to the optimal solution of the Proximal scheme if $\epsilon$ is small. The quantity $\epsilon$ being small implies that the gradients of the subsequent iterations of the Early stopping scheme are close to each other. Secondly, notice that the bound also depends on $k$. We expect the bound to be looser if $k$ is large (everything else remaining the same). In the next section, we investigate this relationship empirically. We observe that for several parameter values of $\alpha, \lambda$ and $k$ satisfying $\lambda \alpha k = 1$, ES and {\sc Prox} methods obtain similar solutions. We show that in those cases, the gradient differences are small. We also show cases where the solutions of {\sc Prox} and those of ES are not close to each other, and argue how in those cases the bound from Theorem 1 is weaker.

\begin{figure}
\includegraphics[width = 0.5\textwidth]{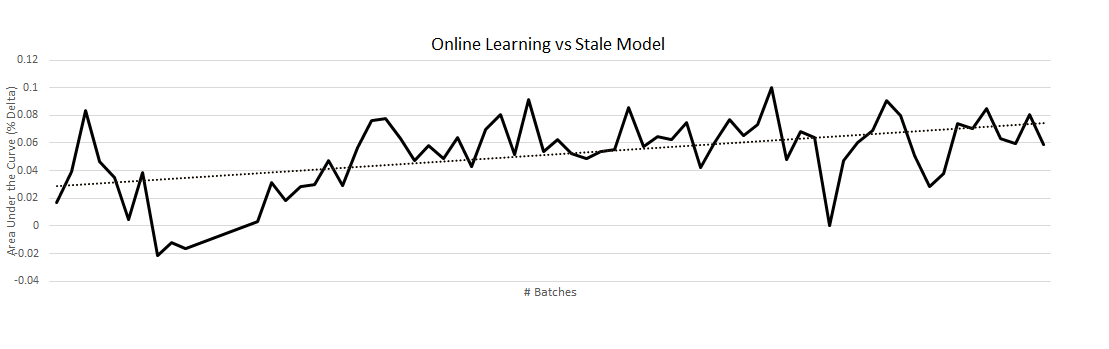}
\includegraphics[width = 0.5\textwidth]{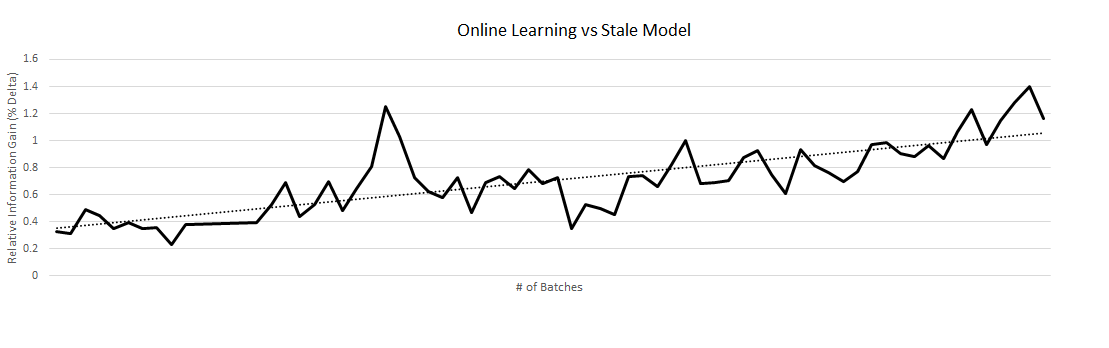}
\caption{Gains from Online Learning relative to a Stale (fixed) Model over a span of three months}
\label{OLgains}
\end{figure}

\begin{figure} 
\arxiv{\includegraphics[width = 0.5\textwidth]{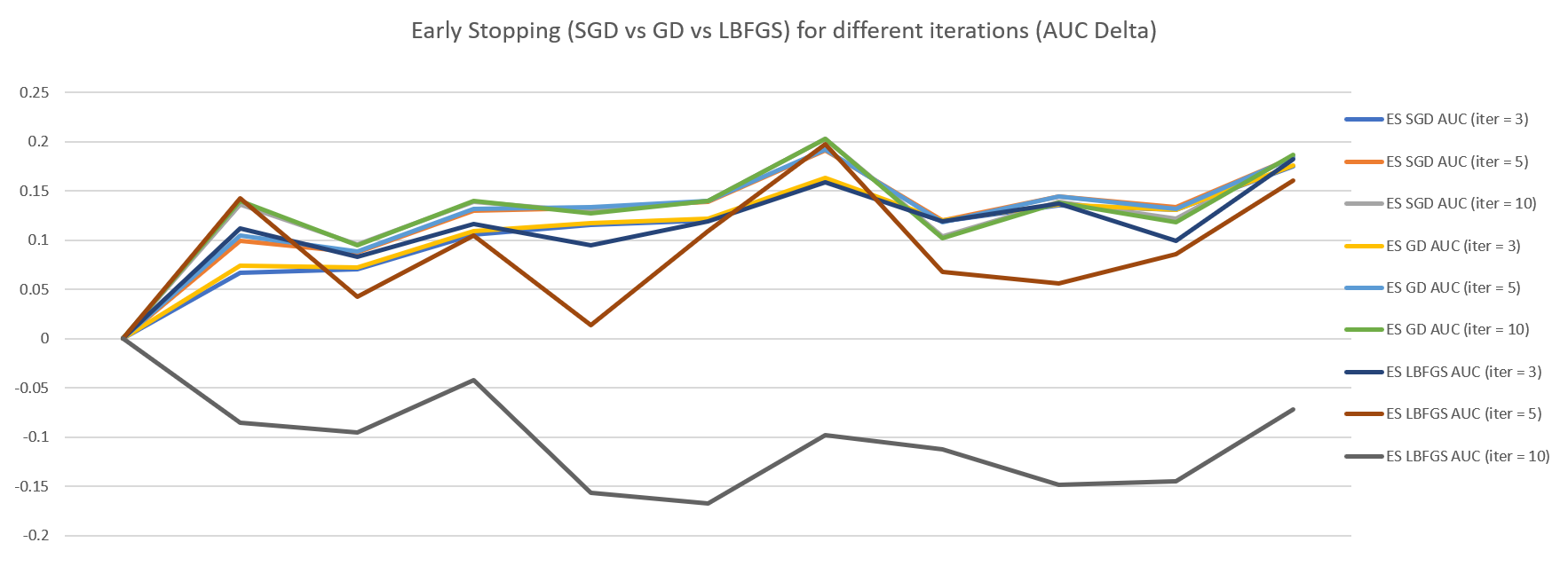}
\includegraphics[width = 0.5\textwidth]{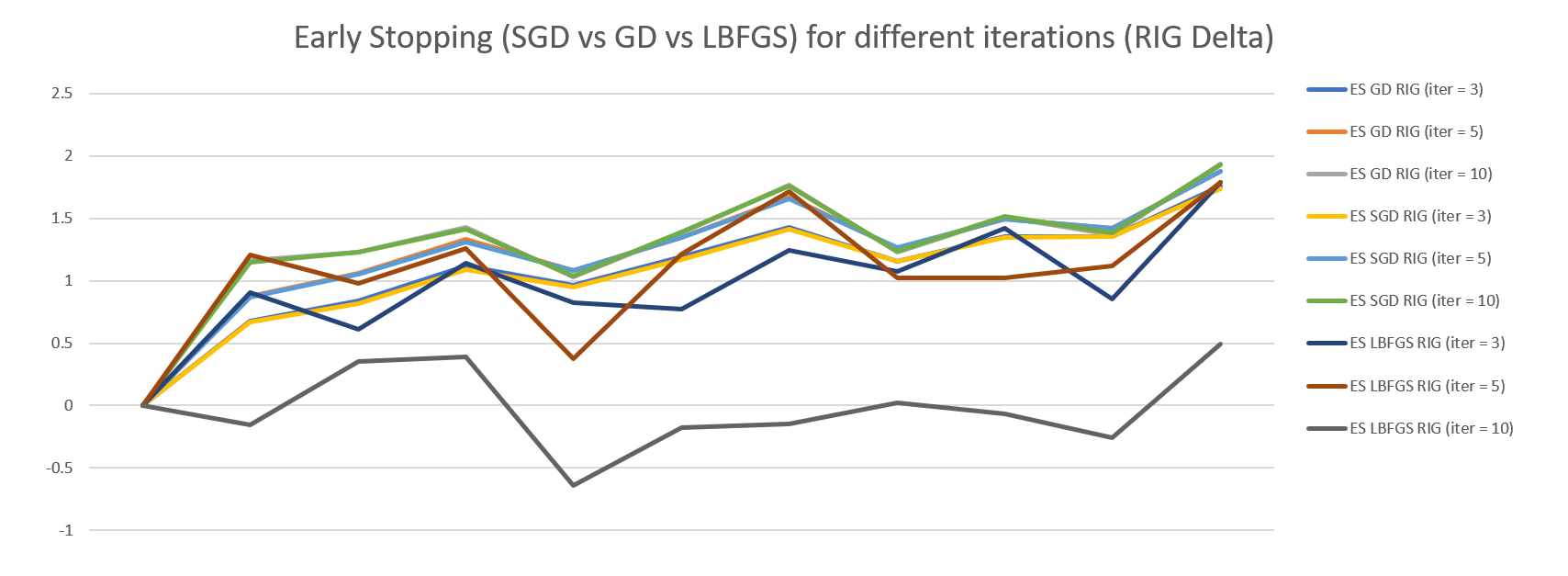}}
\includegraphics[width = 0.5\textwidth]{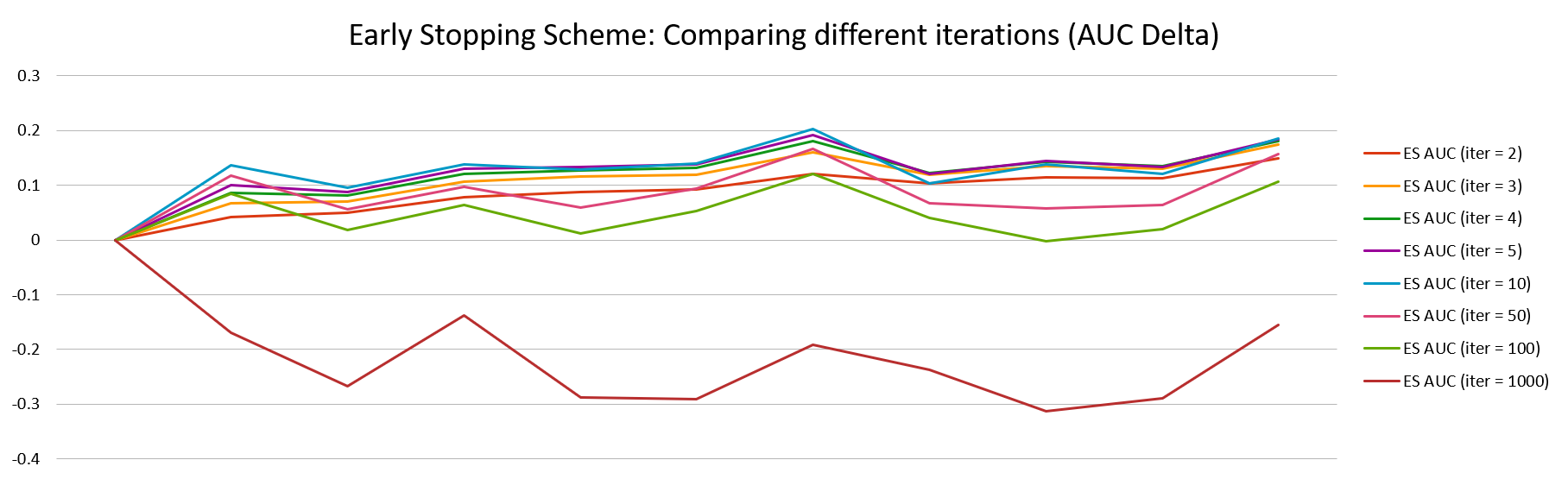}
\includegraphics[width = 0.5\textwidth]{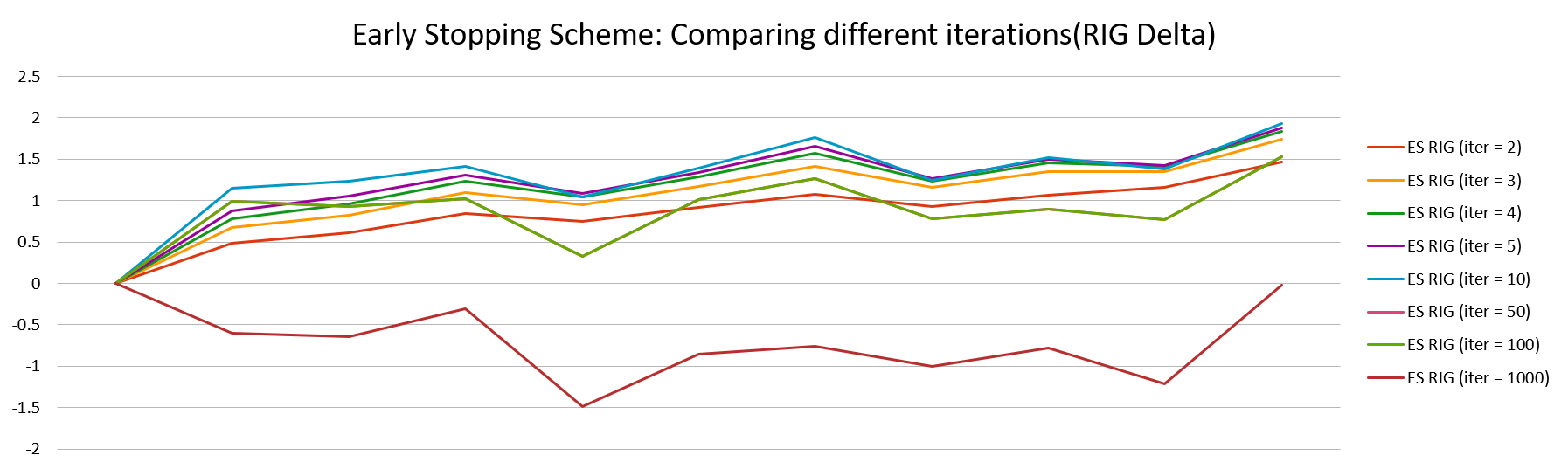}
\includegraphics[width = 0.5\textwidth]{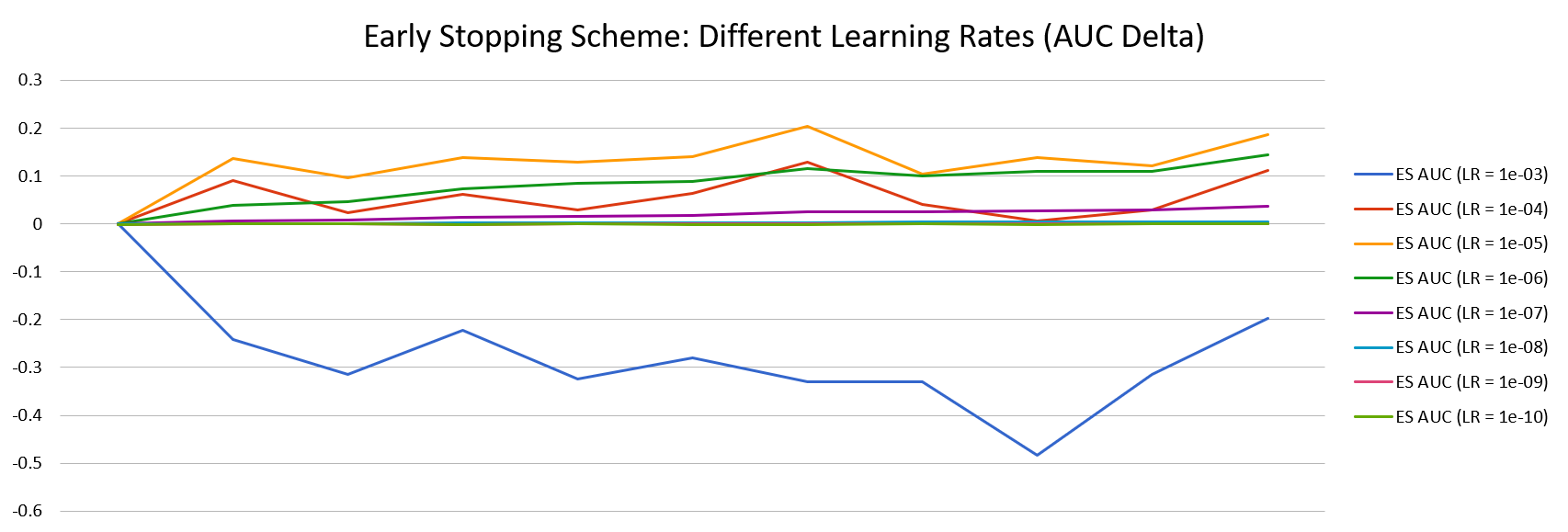}
\includegraphics[width = 0.5\textwidth]{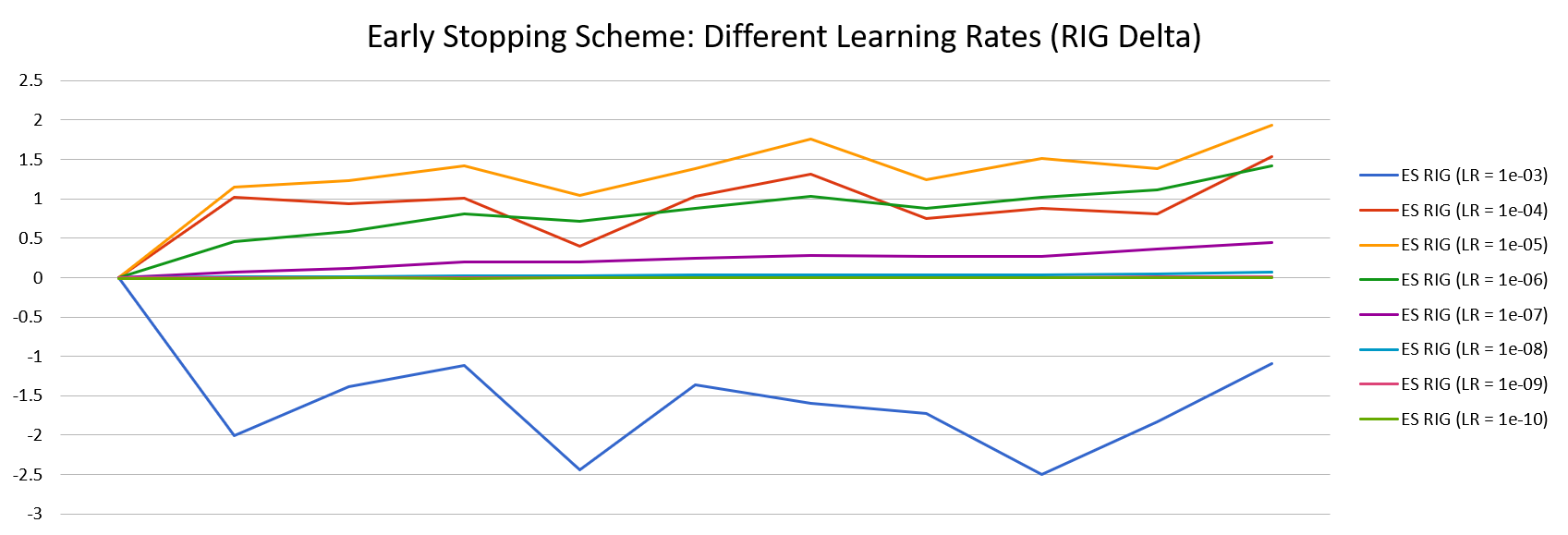}
\caption{\arxiv{The top two figures compares the different algorithms for ES schemes. The third and fourth figure (from top) show the AUC and RIG gains for different number of iterations ($k$) while the bottom two figures show the AUC and RIG gains respectively of the different learning rate parameters ($\alpha$).}\narxiv{The top two figures (from top) show the AUC and RIG gains for different number of iterations ($k$) while the bottom two figures show the AUC and RIG gains respectively of the different learning rate parameters ($\alpha$).}}
\label{escomparisons}
\end{figure}

\begin{figure} 
\includegraphics[width = 0.5\textwidth]{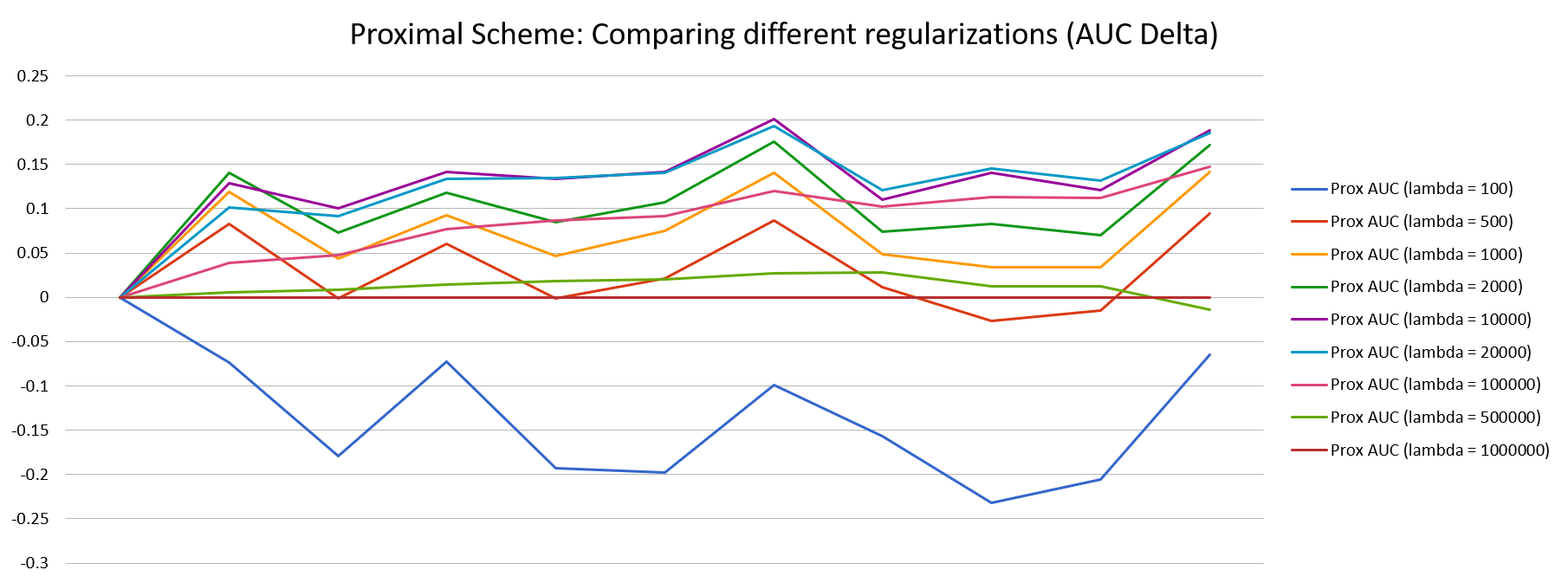}
\includegraphics[width = 0.5\textwidth]{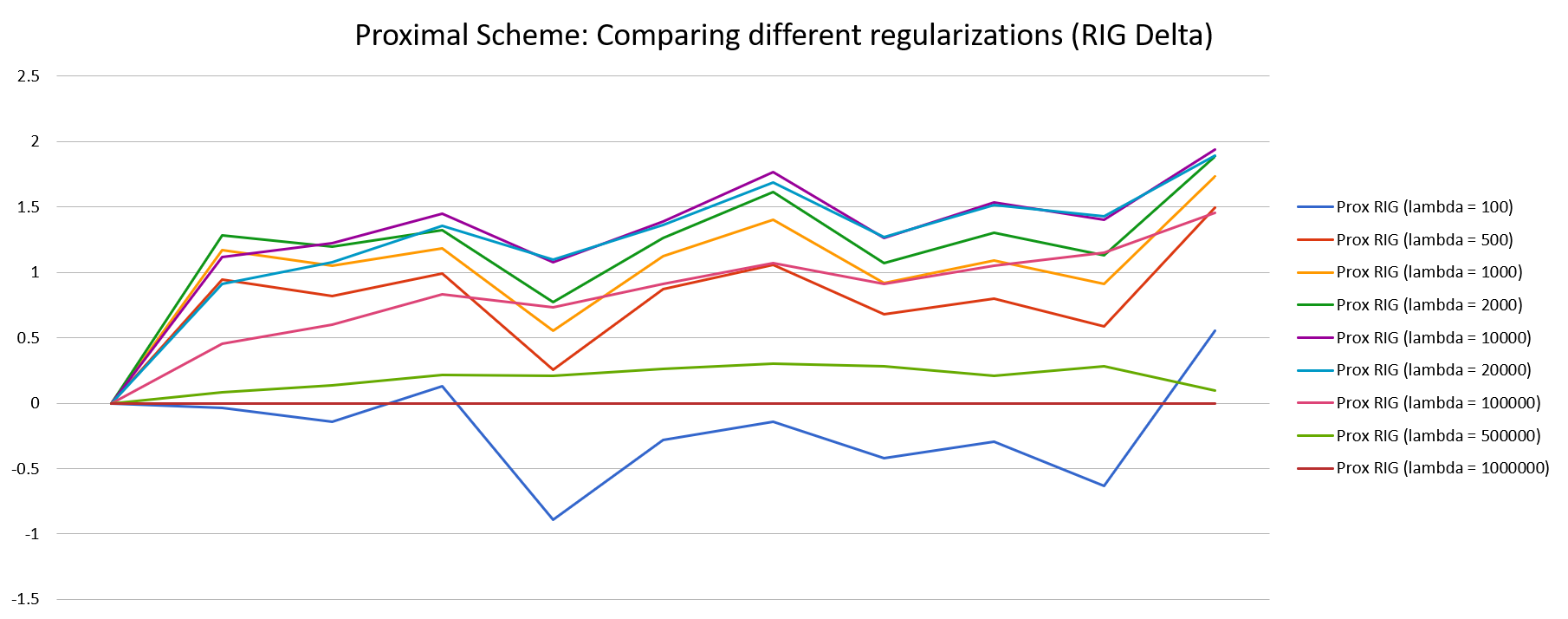}
\caption{The top figure compares the AUC gains and the second figure compares the RIG gains of the Proximal scheme for different values of the proximal regularization $\lambda$.}
\label{proxcomparisons}
\end{figure}

\begin{figure} 
\includegraphics[width = 0.5\textwidth]{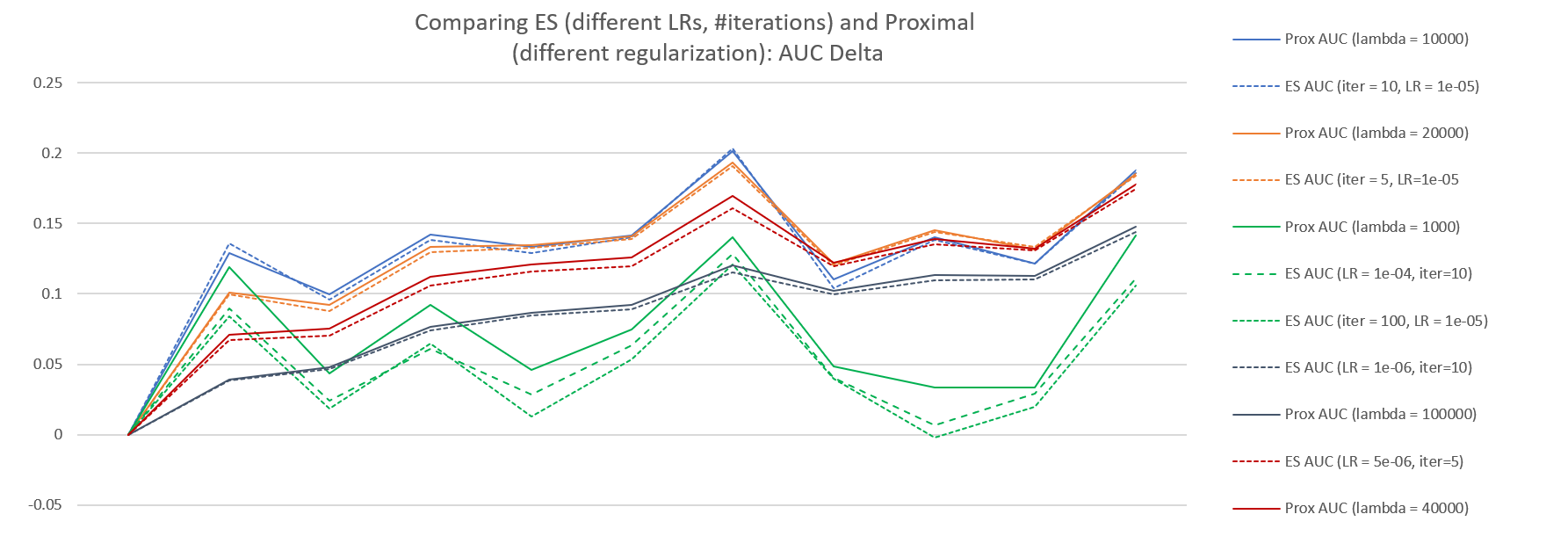}
\includegraphics[width = 0.5\textwidth]{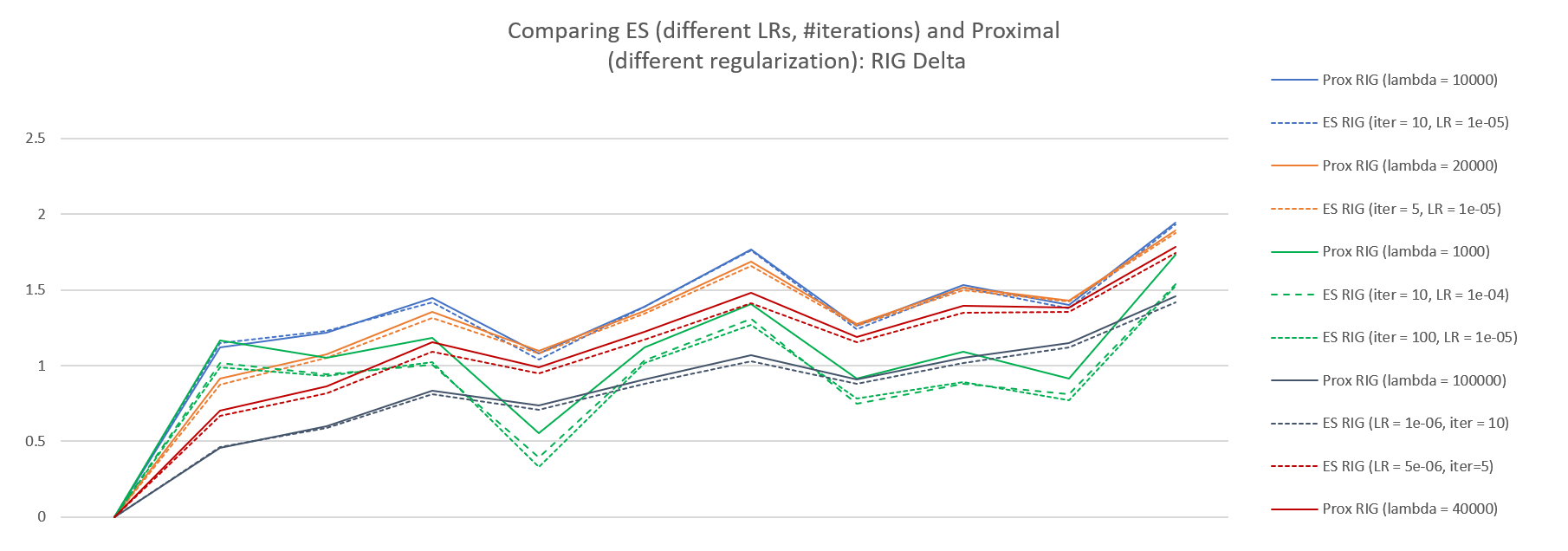}
\caption{Comparing the Proximal Update algorithm and the early stopping algorithms over different number of iterations, learning rates and regularization.}
\label{proxes}
\end{figure}

\begin{figure}
\includegraphics[width = 0.5\textwidth]{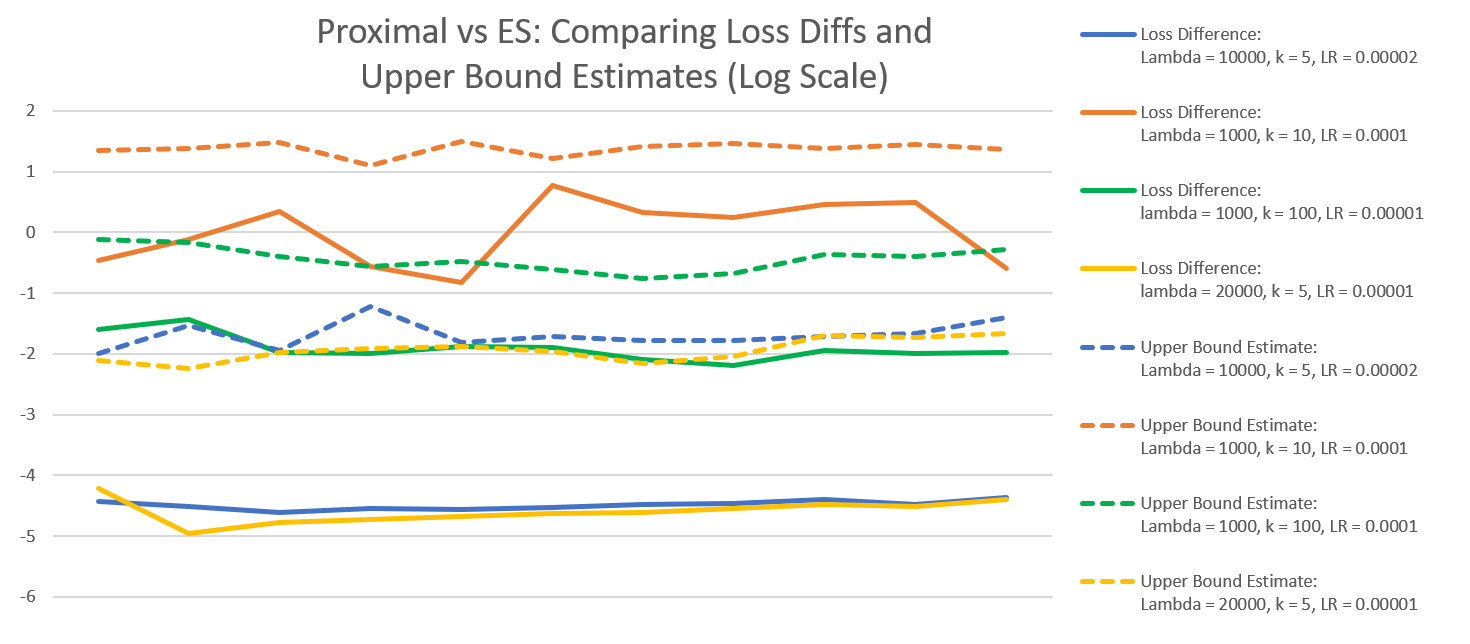}
\caption{Comparing the Difference between the ES and the {\sc Prox} solutions $|G(w_k) - G(w^*)$ and the upper bound from Theorem 1 for different values of $\alpha, k$ and $\lambda$. Results are in Log-scale.}
\label{proxesdiff}
\end{figure}

\begin{figure}
\includegraphics[width = 0.5\textwidth]{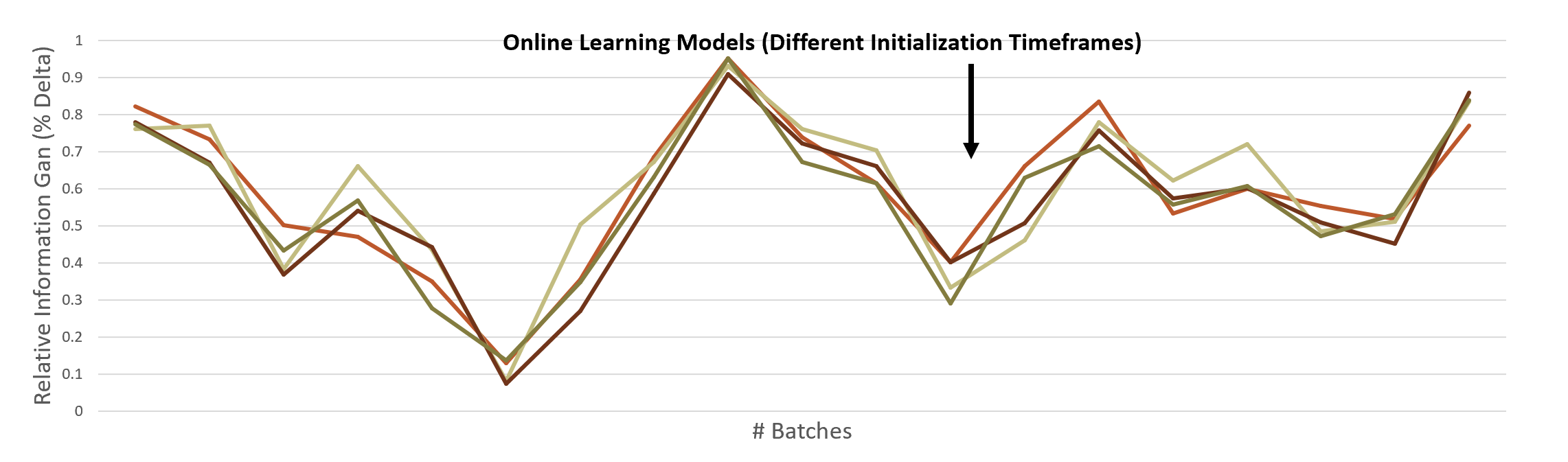}
\caption{Demonstrating the effect of starting OL with different initializations.}
\label{diffinitial}
\end{figure}

\begin{figure}
\includegraphics[width = 0.5\textwidth]{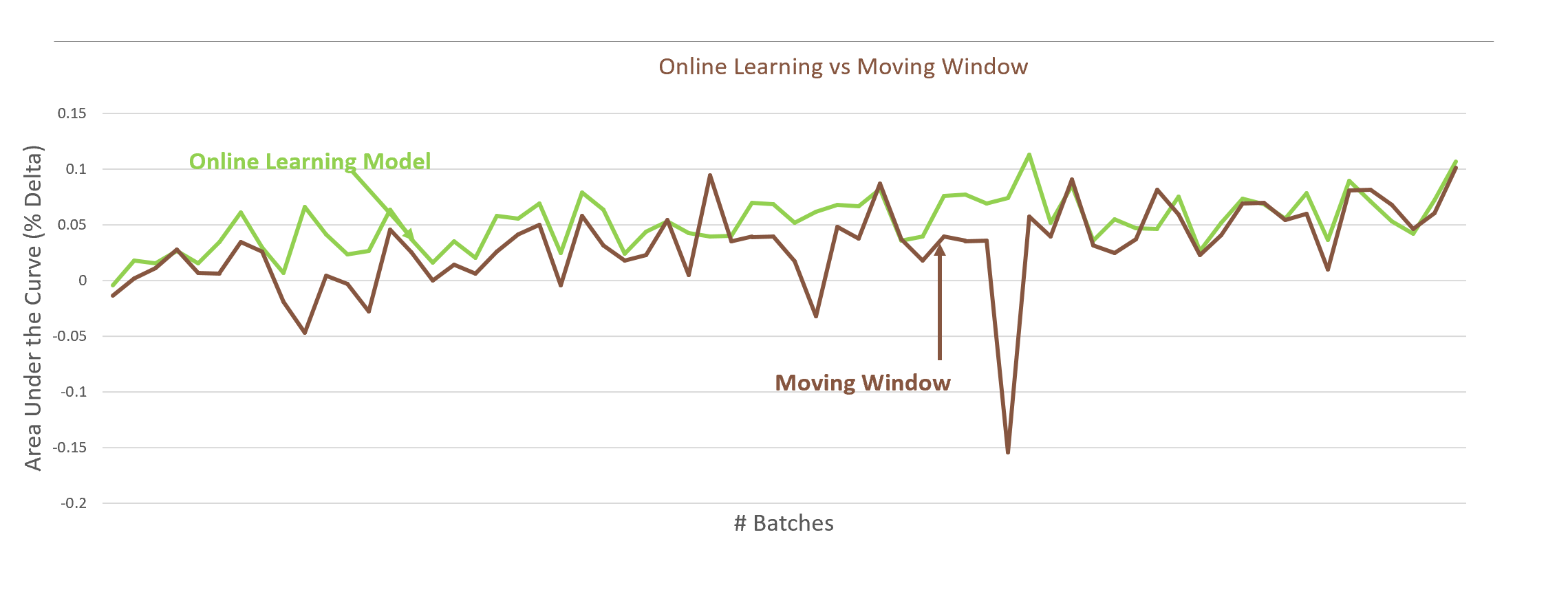}
\includegraphics[width = 0.5\textwidth]{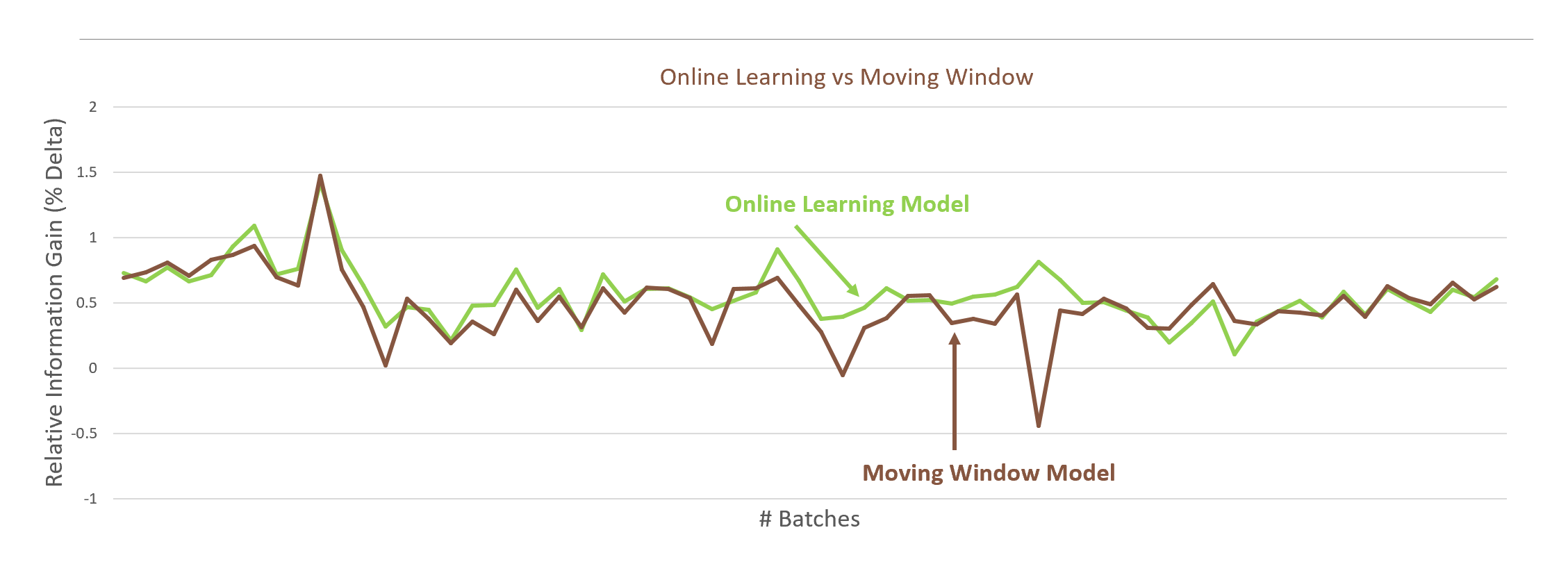}
\caption{The two figure compare Online Learning and the Moving window baseline relative to the base (stale) model on both AUC and RIG respectively.}
\label{OLvsMW}
\end{figure}

\begin{figure}
\includegraphics[width = 0.5\textwidth]{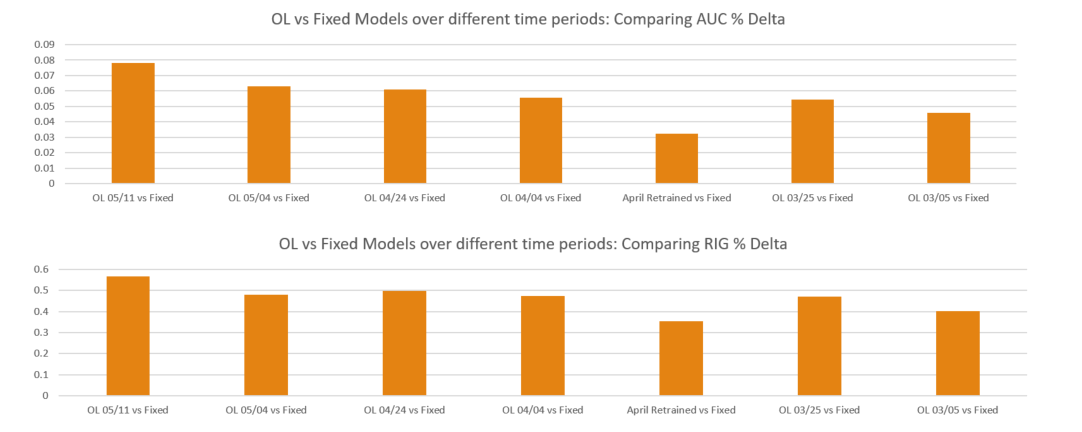}
\caption{Comparison of different OL models having predictions delayed by different time periods.}
\label{delay}
\end{figure}

\section{Experiments and Results}
This section provides details of our extensive evaluation of our online learning framework, with a goal of providing a better understanding to the model performance in various scenarios, and to understand the theoretical results discussed above. The experiments shared below have been run for over a year in our production systems. We have evaluated the models on various feature sets, various times of the year (holiday and regular time periods), and various parameter choices. The model performance is consistent over all these experiments. In the interest of space, we provide only a summary of the results below. The results do not drastically change with different batch sizes (daily, four hourly etc.) Smaller batch sizes only ensure quicker model updates. All our results are on batch sizes of one day. Also, all the results below were conducted over a span of 15 days to three months with daily updates. Each day of data consists of around 2 Million instances. We show the results as time series graphs to demonstrate the gains of online learning over time. Our C++ code (built on top of ~\cite{iyer2018jensen}) and dataset used for our experiments is available at \url{https://github.com/rishabhk108/jensen-ol}.

\subsection{Batch OL as a solution for the Model Staleness}
Figure~\ref{OLgains} demonstrate the gains of online learning by comparing the model metrics to the stale model. We show the gains in both AUC (ranking) and RIG. We see that
the relative RIG gains of about
0.5\% in the middle and 
close to 1.5\% towards the end.  We also observe AUC
gains of around 0.1\%. These experiments are run over a span of three months. Both these are significant gains in our system, and are better than the gains we would expect from retraining the models (we shall compare both in later sections). 

\subsection{Tradeoff parameters for Early Stopping Online Learning}
This section investigates the critical trade-off parameters for Early stopping, namely the choice of the ES algorithm, the Learning rate ($\alpha)$ and the number of iterations ($k$). Figure~\ref{escomparisons} show the results. 

\arxiv{We first compare the different ES algorithms. In this setup, we compare LBFGS, SGD and Gradient Descent. SGD and GD are gradient descent style algorithms and in both cases, we use a fixed learning rate, wherever applicable, and compare their performance for varying numbers of iterations. LBFGS adapts the learning rate as the algorithm proceeds. We see that incremental training with GD and SGD perform similar to one another for the same learning rate and number of iterations -- we run both algorithms with $\alpha = 1e-05$ and obtain results for $k = 3, 5$ and $10$. LBFGS, however, performs worse than both these (with $k = 10$ we see that LBFGS already overfits to the new data). The added benefit of SGD and GD comes from the flexibility of a fixed learning rate, whereas LBFGS tries to minimize the objective function completely as quickly as possible. In our case, we do not want to overfit to the new data, and it is desirable to have the right knobs to tradeoff between the historical and new data. This consideration does not favor LBFGS as the algorithm for use in an ES scheme. This comparison is shown in the top two graphs in Figure~\ref{escomparisons}.}

We \arxiv{next}\narxiv{first} compare the different number of iterations. We set the ES algorithm to be SGD and fix the learning rate as $\alpha = 1e-05$. With a small number of iterations ($k = 2, 3$), the model does not learn enough of the new data while with large number of iterations ($k = 50, 100, 1000$), the model overfits to the new data and we see a loss in performance. The optimal performance is achieved for $k = 5, 10$. This parameter, along with the learning rate, needs to be tuned for each model, depending on the amount of historical data and incremental data. We see the RIG and AUC gains in the \arxiv{third and the fourth graphs (from top)}\narxiv{top two graphs} in Figure~\ref{escomparisons}.

Finally, we compare the learning rate. A large learning rate ($\alpha = 1e-02, 1e-03$), causes the weights to diverge while with a small learning rate ($\alpha = 1e-07, 1e-08$), the learning is slow. We achieve the best results with $\alpha = 1e-05$. The RIG and AUC gains are shown in the last two plots (from top) in Figure~\ref{escomparisons}. \narxiv{In the interest of space, the comparisons of the different early stopping algorithms is in the extended version.}

\subsection{Trade-off Parameters for Proximal Scheme}
We next compare the effect of different regularization parameters for {\sc Prox}. Using a small regularization parameter ($\lambda = 100, 1000$) tends to make the model overfit to the new data, while when using a large regularization $\lambda = 100000, 500000$ and above, the model hardly learns. The optimal performance comes from $\lambda = 10000$ and $\lambda = 20000$ in this case. Again, this parameter will need to be tuned depending on the amount of historical and new data. The results are shown in Figure~\ref{proxcomparisons}.

\subsection{Comparing Early Stopping and Proximal Updates}
In this section, we look into the early stopping and proximal updates, and their connection. The goal of this exercise is to compare several ES schemes (for different $\alpha, k$) and different {\sc Prox} schemes by varying $\lambda$. The results of this are in Figure~\ref{proxes}. Firstly, we compare the following sets of ES and {\sc Prox} schemes, 1) $\alpha = 1e-05, k = 5$ and $\lambda = 20000$, 2) $\alpha = 1e-05, k = 10$ and $\lambda = 10000$, 3) $\alpha = 5e-06, k = 5$ and $\lambda = 40000$, and 4) $\alpha = 1e-06, k = 10$ and $\lambda = 100000$. Notice that all these sets satisfy $\alpha k \lambda = 1$. We see that the {\sc Prox} and {\sc ES} gains are very similar to each other (the blue, orange, red and dark gray lines in Figure~\ref{proxes}). The results hold for both the AUC and RIG gains. We next consider two additional settings: $\alpha = 1e-04, k = 10$ and $\lambda = 1000$ and $\alpha = 1e-05, k = 100, \lambda = 1000$. We see here that there is a gap between the {\sc Prox} and ES schemes with the {\sc Prox} method consistently outperforming the ES schemes in both these cases (the three green lines in Figure~\ref{proxes}).

To understand this better, we plot the difference in loss function $|G(w_k) - G(w^*)|$, and the upper bound from Theorem 1 in Figure~\ref{proxesdiff}. We see that the settings, $\alpha = 1e-05, k = 5, \lambda = 20000$ and $\alpha = 2e-05, k = 5, \lambda = 10000$ have small values of the loss function difference, as expected. We see that the upper bound estimate is also small in this case (around 1e-02). However, with the settings $\alpha = 1e-04, k = 10$ and $\lambda = 1000$ and $\alpha = 1e-05, k = 100, \lambda = 1000$, we see a larger difference between $G(w^*)$ and $G(w_k)$ (i.e. the {\sc Prox} and the {\sc ES} solutions). Note that all four of these satisfy $\alpha k \lambda = 1$. We also see that the upper bound estimate is also larger. With a larger learning rate $\alpha = 1e-04$, the gradient difference between subsequent iterations will be larger, so will $\epsilon$. In the second case, the learning rate is smaller $\alpha = 1e-05$, but we run it for more iterations. Correspondingly, the bound (which depends on both $\epsilon$ and $k$) is larger. 


{\sc ES} and {\sc Prox} yield very similar update rules if we choose the right set of hyper-parameters. Fortunately, in practice, we observe that the the optimal performance comes from smaller values of $\alpha$ and $k$, and in those settings the {\sc Prox} and the {\sc ES} schemes coincide. Unlike the {\sc Prox} scheme, the {\sc ES} stopping does not require solving a convex optimization scheme to completion. We just need to run a few iterations of SGD with the right learning rate. On the flip side, ES scheme has more hyper parameters ($\alpha, k$) which make it slightly harder to tune compared to {\sc Prox}, where we just need to tune the regularization. In the rest of the paper, we choose the setting with $\alpha = 1e-05$ and $k = 10$ and use the {\sc ES} update scheme.

\subsection{Comparing OL with Regular Model Retrains}
We have illustrated how batch OL can help resolve the issue of model staleness that a fixed model suffers from. Another obvious alternative to resolve this problem is to automatically retrain and update the base model periodically. To compare these 2 approaches, we setup a baseline where the model was being completed retrained daily on the previous week’s data and evaluated on just the next day. This was compared against a batch OL model being incrementally trained daily. Figure~\ref{OLvsMW} demonstrates the results.

Firstly we notice that both models follow the same trend over time. Next we see that on several days, the online learning outperforms the moving window baseline. This can be attributed to the fact that online learning is done incrementally in each batch, so the model has seen more data than just the previous period. This gives the OL model the ability to generalize better to trends it has learnt in earlier batches, while also having the ability to learn and adapt to the recent trends. The model retraining however learns solely from the last few days and may overfit to this distribution. We also see that the moving window baseline is more sensitive to overfitting to data corruption and distributional changes compared to online learning. The reader will notice large drops in AUC and RIG owing to distributional changes (see the two large srops with the Moving window approach). This effect can be even more pronounced with data corruption issues (like system bugs or livesites). Since online learning adapts slowly while remembering the past historical data, it does not overfit as much. Moreover, it is easier to also implement validation checks and safeguards to ensure it does not learn from corrupt data.

\subsection{Impact of when we start Online Learning}
We next study the effect of online learning on different initializations. We consider different starting points of online learning. In this experiment, we train four different base models with one week of initial data each. We start four different OL schemes, each of which begin one week apart. We see that after about a month of online learning, all the four model converge to roughly the same performance with respect to a fixed base model. Figure~\ref{diffinitial} shows the results of this.

\subsection{Delay Analysis in Online Learning}
In this section, we investigate the effect of delayed predictions made by online learning models. In other words, we fix the model evaluation period and compare the performance of multiple OL models trained till different time periods. The most recent model was trained on daily batches till one day before this period. The other models in comparison are trained till one week, 15 days and so on up till more than 2 months before the evaluation period. We also compare the performance of a base model trained from scratch approximately 1 month before the evaluation period. The results are in Figure~\ref{delay}. Here we can see that for both AUC and RIG, the performance degrades with increased delay. This inference is intuitive since the delayed models haven’t seen the latest trends in data closer to the evaluation period. The reduction in performance however is small as we move to older models. Even the model trained till 03/05, which is more than 2 months before the evaluation period, retains most of the gains in AUC and RIG over the base model.
We also compare these delayed models to a fixed baseline model trained to completion around one month before the evaluation period (marked as April Retrained, trained on 04/01 to 04/07). Notice that there are a few OL model snapshots that have been updated till before this time-period, namely till 03/25 and 03/05. As seen in the figure, even these OL models perform better than the retrained baseline, even though the baseline model is trained closer to the evaluation period. The reason for this is that the OL models are trained incrementally and have actually seen data across several months, hence they generalize better. The fixed model on the other hand trains on just one week of data and stands to learn just from the distribution in this time period. This again underscores the point that online learning models are superior compared to simple retrained models with a data refresh.

\section{Conclusions and Lessons Learned}
This paper presents a unified framework for online learning, by showing how two seemingly different views of online learning, namely iterative early stopping scheme and a Proximal Update algorithm (both of which have been extensively in literature for this problem), are closely related. We provide conditions when the two algorithms achieve the same updates and empirically validate them. We demonstrate several results proving the benefit of online learning, by understanding the tradeoff between historical and new data, the impact of initializations and delay in the system, and proving that Online Learning is a superior and more stable method compared to model retrainings. 

Finally, we discuss some important validation and safeguard mechanisms required for online learning systems in production systems. This is important since models are getting automatically updated. Some of our validation checks include: 
\begin{itemize}
\item Check daily differences in model metrics such as RIG and AUC (day over day differences). We do not expect the day over day differences to be large due to the incremental nature of our online learning schemes.
\item Check differences to the base model. We expect to see non trivial improvements compared to stale initial model.
\item Comparison the moving window ensures that the online learning is no worse than a retrained baseline model.
\item Day over day CTR and other data checks is required to ensure we do not incrementally train models over livesight data. For data checks, we check the volume of the training data over various slices as we do not expect a drastic difference in the volume of the input data.
\end{itemize}
We also have monitoring dashboards to monitor daily model metrics, input data volumes, CTR etc. These dashboards allow us to monitor the daily performance of the models, and investigate potential issues. In case of model issues, it is also easy to rollback to previous snapshots of the model.
\bibliographystyle{aaai}
\bibliography{main}

\begin{thebibliography}{}

\bibitem[\protect\citeauthoryear{Bottou, Curtis, and
  Nocedal}{2018}]{bottou2018optimization}
Bottou, L.; Curtis, F.~E.; and Nocedal, J.
\newblock 2018.
\newblock Optimization methods for large-scale machine learning.
\newblock {\em SIAM Review} 60(2):223--311.

\bibitem[\protect\citeauthoryear{Chapelle and Li}{2011}]{chapelle2011empirical}
Chapelle, O., and Li, L.
\newblock 2011.
\newblock An empirical evaluation of thompson sampling.
\newblock In {\em Advances in neural information processing systems},
  2249--2257.

\bibitem[\protect\citeauthoryear{Chapelle, Manavoglu, and
  Rosales}{2015}]{chapelle2015simple}
Chapelle, O.; Manavoglu, E.; and Rosales, R.
\newblock 2015.
\newblock Simple and scalable response prediction for display advertising.
\newblock {\em ACM Transactions on Intelligent Systems and Technology (TIST)}
  5(4):61.

\bibitem[\protect\citeauthoryear{Cheng and
  Cant{\'u}-Paz}{2010}]{cheng2010personalized}
Cheng, H., and Cant{\'u}-Paz, E.
\newblock 2010.
\newblock Personalized click prediction in sponsored search.
\newblock In {\em Proceedings of the third ACM international conference on Web
  search and data mining},  351--360.
\newblock ACM.

\bibitem[\protect\citeauthoryear{Ciaramita, Murdock, and
  Plachouras}{2008}]{ciaramita2008online}
Ciaramita, M.; Murdock, V.; and Plachouras, V.
\newblock 2008.
\newblock Online learning from click data for sponsored search.
\newblock In {\em Proceedings of the 17th international conference on World
  Wide Web},  227--236.
\newblock ACM.

\bibitem[\protect\citeauthoryear{Duchi, Hazan, and
  Singer}{2011}]{duchi2011adaptive}
Duchi, J.; Hazan, E.; and Singer, Y.
\newblock 2011.
\newblock Adaptive subgradient methods for online learning and stochastic
  optimization.
\newblock {\em Journal of Machine Learning Research} 12(Jul):2121--2159.

\bibitem[\protect\citeauthoryear{Graepel \bgroup et al\mbox.\egroup
  }{2010}]{graepel2010web}
Graepel, T.; Candela, J.~Q.; Borchert, T.; and Herbrich, R.
\newblock 2010.
\newblock Web-scale bayesian click-through rate prediction for sponsored search
  advertising in microsoft's bing search engine.
\newblock Omnipress.

\bibitem[\protect\citeauthoryear{He \bgroup et al\mbox.\egroup
  }{2014}]{he2014practical}
He, X.; Pan, J.; Jin, O.; Xu, T.; Liu, B.; Xu, T.; Shi, Y.; Atallah, A.;
  Herbrich, R.; Bowers, S.; et~al.
\newblock 2014.
\newblock Practical lessons from predicting clicks on ads at facebook.
\newblock In {\em Proceedings of the Eighth International Workshop on Data
  Mining for Online Advertising},  1--9.
\newblock ACM.

\bibitem[\protect\citeauthoryear{Iyer, Halloran, and
  Wei}{2018}]{iyer2018jensen}
Iyer, R.; Halloran, J.~T.; and Wei, K.
\newblock 2018.
\newblock Jensen: An easily-extensible c++ toolkit for production-level machine
  learning and convex optimization.
\newblock {\em arXiv preprint arXiv:1807.06574}.

\bibitem[\protect\citeauthoryear{Kirkpatrick \bgroup et al\mbox.\egroup
  }{}]{kirkpatrick2017overcoming}
Kirkpatrick, J.; Pascanu, R.; Rabinowitz, N.; Veness, J.; Desjardins, G.; Rusu,
  A.~A.; Milan, K.; Quan, J.; Ramalho, T.; Grabska-Barwinska, A.; et~al.
\newblock Overcoming catastrophic forgetting in neural networks.
\newblock {\em Proceedings of the national academy of sciences}.

\bibitem[\protect\citeauthoryear{Lin, Weng, and Keerthi}{2007}]{lin2007trust}
Lin, C.-J.; Weng, R.~C.; and Keerthi, S.~S.
\newblock 2007.
\newblock Trust region newton methods for large-scale logistic regression.
\newblock In {\em Proceedings of the 24th international conference on Machine
  learning},  561--568.
\newblock ACM.

\bibitem[\protect\citeauthoryear{Ling \bgroup et al\mbox.\egroup
  }{2017}]{ling2017model}
Ling, X.; Deng, W.; Gu, C.; Zhou, H.; Li, C.; and Sun, F.
\newblock 2017.
\newblock Model ensemble for click prediction in bing search ads.
\newblock In {\em Proceedings of the 26th International Conference on World
  Wide Web Companion},  689--698.
\newblock International World Wide Web Conferences Steering Committee.

\bibitem[\protect\citeauthoryear{Liu and Nocedal}{1989}]{liu1989limited}
Liu, D.~C., and Nocedal, J.
\newblock 1989.
\newblock On the limited memory bfgs method for large scale optimization.
\newblock {\em Mathematical programming} 45(1-3):503--528.

\bibitem[\protect\citeauthoryear{Liu \bgroup et al\mbox.\egroup
  }{2017}]{liu2017pbodl}
Liu, X.; Xue, W.; Xiao, L.; and Zhang, B.
\newblock 2017.
\newblock Pbodl: Parallel bayesian online deep learning for click-through rate
  prediction in tencent advertising system.
\newblock {\em arXiv preprint arXiv:1707.00802}.

\bibitem[\protect\citeauthoryear{Ma \bgroup et al\mbox.\egroup
  }{2009}]{ma2009identifying}
Ma, J.; Saul, L.~K.; Savage, S.; and Voelker, G.~M.
\newblock 2009.
\newblock Identifying suspicious urls: an application of large-scale online
  learning.
\newblock In {\em Proceedings of the 26th annual international conference on
  machine learning},  681--688.
\newblock ACM.

\bibitem[\protect\citeauthoryear{McMahan and Streeter}{2014}]{mcmahan2014delay}
McMahan, B., and Streeter, M.
\newblock 2014.
\newblock Delay-tolerant algorithms for asynchronous distributed online
  learning.
\newblock In {\em Advances in Neural Information Processing Systems},
  2915--2923.

\bibitem[\protect\citeauthoryear{McMahan \bgroup et al\mbox.\egroup
  }{2013}]{mcmahan2013ad}
McMahan, H.~B.; Holt, G.; Sculley, D.; Young, M.; Ebner, D.; Grady, J.; Nie,
  L.; Phillips, T.; Davydov, E.; Golovin, D.; et~al.
\newblock 2013.
\newblock Ad click prediction: a view from the trenches.
\newblock In {\em Proceedings of the 19th ACM SIGKDD international conference
  on Knowledge discovery and data mining},  1222--1230.
\newblock ACM.

\bibitem[\protect\citeauthoryear{McMahan}{2011}]{mcmahan2011follow}
McMahan, B.
\newblock 2011.
\newblock Follow-the-regularized-leader and mirror descent: Equivalence
  theorems and l1 regularization.
\newblock In {\em Proceedings of the Fourteenth International Conference on
  Artificial Intelligence and Statistics},  525--533.

\bibitem[\protect\citeauthoryear{Shalev-Shwartz and
  others}{2012}]{shalev2012online}
Shalev-Shwartz, S., et~al.
\newblock 2012.
\newblock Online learning and online convex optimization.
\newblock {\em Foundations and Trends{\textregistered} in Machine Learning}
  4(2):107--194.

\bibitem[\protect\citeauthoryear{Shalev-Shwartz and
  Singer}{2007}]{shalev2007online}
Shalev-Shwartz, S., and Singer, Y.
\newblock 2007.
\newblock Online learning: Theory, algorithms, and applications.

\bibitem[\protect\citeauthoryear{Zinkevich}{2003}]{zinkevich2003online}
Zinkevich, M.
\newblock 2003.
\newblock Online convex programming and generalized infinitesimal gradient
  ascent.
\newblock In {\em Proceedings of the 20th International Conference on Machine
  Learning (ICML-03)},  928--936.

\end{thebibliography}
\arxiv{
\section{Appendix}
Proof of Theorem 1.
\begin{proof}
To prove this result, we note that the ES produces a set of weights $w_0, w_1, \cdots, w_k$. Given that we run ES with a gradient descent with a fixed learning rate $\alpha$, it is easy to see that $w_i = w_0 - \sum_{j = 0}^{i-1} \alpha \nabla F(w_j)$. Also note that $\nabla G(w_k) = \nabla F(w_k) + \lambda (w_k - w_0)$. From the above, $w_k - w_0 = -\sum_{j = 0}^{k-1} \nabla F(w_j)$. Therefore,
\begin{align}
\nabla G(w_k) = \nabla F(w_k) - \lambda \alpha \sum_{j = 1}^{k-1} \nabla F(w_j).
\end{align}
If $\lambda \alpha k = 1$, then we have,
\begin{align}
\nabla G(w_k) &= \nabla F(w_k) - \sum_{j = 0}^{k-1} \nabla F(w_j)/k \\
		&= \sum_{j = 0}^{k-1} (\nabla F(w_k) - \nabla F(w_j))/k
\end{align}
It then follows that 
\begin{align}
||\nabla G(w_k) || \leq  \sum_{j = 1}^{k-1} ||\nabla F(w_k) - \nabla F(w_j)||.
\end{align}
Given that $\epsilon = \max_i || \nabla F(w_i) - \nabla F(w_{i-1}) ||$, we can show that $||\nabla F(w_k) - \nabla F(w_j)|| \leq (k-j)\epsilon$. Adding this up, we get that $||\nabla G(w_k) || \leq (k-1)\epsilon$. Since $G$ is convex, we have that,
\begin{align}
G(w_k) \leq G(w^*) + \langle \nabla G(w_k),(w_k - w^*)\rangle
\end{align}
From which, we have $|G(w_k) - G(w^*)| \leq \langle \nabla G(w_k),(w_k - w^*)\rangle \leq ||\nabla G(w_k) || || w_k - w^*|| \leq \epsilon (k - 1) ||w_k - w^*||$.
\end{proof}
The proof of Corollary 1 follows exactly as above, except that we consider a co-ordinate wise sum for each of the expressions.
}

\end{document}